\documentclass[conference]{IEEEtran}
\IEEEoverridecommandlockouts
\usepackage{cite}
\usepackage{amsmath,amssymb,amsfonts}
\usepackage{graphicx}
\usepackage{textcomp}
\usepackage{xcolor}
\usepackage{hyperref}
\usepackage[ruled,vlined,linesnumbered]{algorithm2e}
\usepackage{tabularx}
\usepackage{booktabs}
\usepackage{multirow}
\ifCLASSOPTIONcompsoc
    \usepackage[caption=false, font=normalsize, labelfont=sf, textfont=sf]{subfig}
\else
\usepackage[caption=false, font=footnotesize]{subfig}

\usepackage{cleveref}
\def\BibTeX{{\rm B\kern-.05em{\sc i\kern-.025em b}\kern-.08em
    T\kern-.1667em\lower.7ex\hbox{E}\kern-.125emX}}
\begin{document}

\title{Dynamic Gated Cross-Modal Fusion with Sarcastic-aware Contrastive Regularization for Multimodal Sarcasm Detection\\
\thanks{\IEEEauthorrefmark{2}Subin Huang and Chao Kong are with Bigdata and Responsible Artificial Intelligence for National Governance, Renmin University of China, Beijing 100872, China.}
\thanks{This work was also supported in part by the Anhui Provincial Natural Science Foundation under Grant No. 2308085MF220.}
}

\author{\IEEEauthorblockN{1\textsuperscript{st} Hao Guo}
\IEEEauthorblockA{\textit{School of Computer and Information} \\
\textit{Anhui Polytechnic University}\\
Wuhu, China \\
\href{mailto:sifan10077@gmail.com}{sifan10077@gmail.com}}
\and
\IEEEauthorblockN{2\textsuperscript{nd} Subin Huang\IEEEauthorrefmark{1}\IEEEauthorrefmark{2} \thanks{* Subin Huang is the corresponding author.}}

\IEEEauthorblockA{\textit{School of Computer and Information} \\
\textit{Anhui Polytechnic University}\\
Wuhu, China \\
\href{mailto:subinhuang@ahpu.edu.cn}{subinhuang@ahpu.edu.cn}
}
\and
\IEEEauthorblockN{3\textsuperscript{rd} Junjie Chen}
\IEEEauthorblockA{\textit{School of Computer and Information} \\
\textit{Anhui Polytechnic University}\\
Wuhu, China \\
\href{mailto:jorji.chen@gmail.com}{jorji.chen@gmail.com}
}
\and
\IEEEauthorblockN{4\textsuperscript{th} Zhifa Geng}
\IEEEauthorblockA{\textit{School of Computer and Information} \\
\textit{Anhui Polytechnic University}\\
Wuhu, China \\
\href{mailto:zhifageng77@gmail.com}{zhifageng77@gmail.com}
}
\and
\IEEEauthorblockN{5\textsuperscript{th} Sanmin Liu}
\IEEEauthorblockA{\textit{School of Computer and Information} \\
\textit{Anhui Polytechnic University}\\
Wuhu, China \\
\href{mailto:sanmin.liu@ahpu.edu.cn}{sanmin.liu@ahpu.edu.cn}
}
\and
\IEEEauthorblockN{6\textsuperscript{th} Chao Kong\IEEEauthorrefmark{2}}
\IEEEauthorblockA{\textit{School of Computer and Information} \\
\textit{Anhui Polytechnic University}\\
Wuhu, China \\
\href{mailto:kongchao@ahpu.edu.cn}{kongchao@ahpu.edu.cn}
}
}

\maketitle

\begin{abstract}
Multimodal sarcasm detection aims to identify sarcastic intent from multimodal content, where inconsistencies between literal meaning and contextual cues often signal irony. This task has attracted increasing research attention. However, accurate detection remains challenging due to instance-dependent modality contributions and misleading semantic consistency, where surface-level alignment masks underlying contradictory intent. Existing methods often rely on fixed fusion strategies and treat sarcasm as generic cross-modal mismatch, limiting their ability to capture subtle sarcasm cues and instance-specific modality interactions.
To address these challenges, we propose a novel MSD framework that integrates Dynamic Gated Cross-Modal Fusion with Sarcastic-aware Contrastive Regularization (SaCR). Specifically, a bidirectional gated interaction module performs cross-modal feature filtering and adaptively calibrates textual and visual contributions at the instance level. A dynamic fusion gate further balances modality importance to generate more robust multimodal representations. Furthermore, SaCR is introduced as a label-aware contrastive regularization objective that encourages semantic consistency for non-sarcastic samples while suppressing misleading consistency in sarcastic cases.
The proposed framework is trained end-to-end with a multi-objective learning strategy that jointly optimizes multimodal classification and auxiliary unimodal supervision. Extensive experiments on MMSD and MMSD2.0 demonstrate that the proposed method consistently outperforms strong baselines.
\end{abstract}


\vspace{-3pt}

\section{Introduction}
\label{sec:intro}

Multimodal sarcasm detection (MSD) aims to identify sarcastic intent from paired textual and visual content~\cite{Schifanella-2016-DetectingSarcasmMultimodal,Cai-2019-MultiModalSarcasmTwitter,Zhuang-2025-MultiModalSarcasm,wang-2026-XInsight}. Unlike general multimodal classification, MSD is particularly challenging because sarcastic evidence is often both instance-dependent and deceptively aligned across modalities~\cite{Qin-2023-MMSD2,Tian-2023-DynamicRoutingTransformer,Qiao-2023-MutualEnhancedIncongruity}. In some cases, sarcasm is mainly triggered by visual contradiction, whereas in others it relies more on textual framing or subtle cross-modal interplay. Moreover, sarcastic image-text pairs may appear semantically compatible at the literal level while conveying contradictory underlying intent, which makes naive alignment-based modeling unreliable~\cite{Li-2025-CommunicationStyleInteraction,Zhu-2024-TFCDCounterfactualDebiasing}.
\begin{figure}
    \centering
    \includegraphics[width=0.75\linewidth]{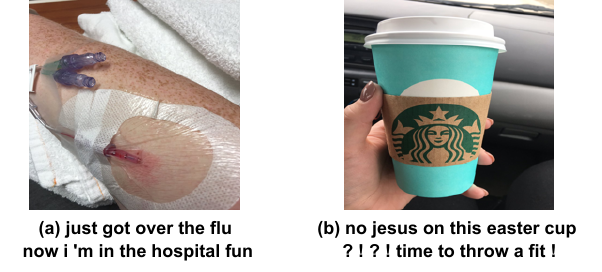}
    \caption{Illustrative examples of surface-level consistency masking underlying incongruity in multimodal sarcasm. (a) The text "fun" and the hospital setting are semantically compatible but sentimentally contradictory. (b) The text "no Jesus" and the cup image share a literal subject, yet convey ironic intent. Our method aims to capture such hidden incongruity beyond literal alignment.}
    \label{fig:introduction}
\end{figure}

Recent MSD research has progressed from early multimodal fusion and incongruity modeling to graph-based interaction, debiasing strategies, and large vision-language model based approaches~\cite{Cai-2019-MultiModalSarcasmTwitter,Xu-2020-ReasoningMultimodalSarcastic,Liang-2021-MultiModalGraphs,Jia-2024-DebiasingMultimodalSarcasm,Li-2025-CommunicationStyleInteraction,Chen-2025-SeeingSarcasmDifferentEyes}. However, existing methods still suffer from two limitations. First, many approaches rely on fixed or insufficiently adaptive fusion schemes, which may dilute the dominant sarcastic cues in a given instance and fail to capture instance-specific modality contributions~\cite{Liang-2022-MultiModalCrossModalGCN,Qin-2023-MMSD2,Tian-2023-DynamicRoutingTransformer}. Second, they often treat sarcasm as generic cross-modal mismatch, without explicitly modeling a common but underexplored phenomenon in sarcasm: misleading semantic consistency that masks underlying intent contradiction~\cite{Qiao-2023-MutualEnhancedIncongruity,Zhu-2024-TFCDCounterfactualDebiasing,Li-2025-CommunicationStyleInteraction}. As a result, models can overemphasize superficial alignment while overlooking the actual evidence of irony.

As illustrated in Figure~\ref{fig:introduction}, the sarcastic signal may originate from different modalities. In Figure~\ref{fig:introduction}(a), the textual cue “fun” contradicts the hospital scene, resulting in text-dominant sarcasm. In contrast, Figure~\ref{fig:introduction}(b) shows a case where sarcasm emerges from visual and contextual interpretation despite surface-level semantic consistency. These examples highlight that sarcasm is both modality-dependent and often masked by superficial alignment, posing significant challenges for multimodal modeling.

To address these challenges, we propose a unified framework that integrates Dynamic Gated Cross-Modal Fusion with Sarcastic-aware Contrastive regularization (SaCR). The former performs bidirectional text-image interaction and adaptively calibrates modality contributions for each instance, while the latter imposes label-aware constraints on cross-modal similarity to distinguish genuine semantic alignment from sarcasm-induced misleading consistency. The whole framework is trained end-to-end under a unified learning objective.

The main contributions of this paper are summarized as follows:
\begin{itemize}
    \item We propose a multimodal sarcasm detection framework with dynamic gated cross-modal fusion, which adaptively balances textual and visual evidence in an instance-aware manner.
    
    \item We introduce a sarcastic-aware contrastive regularization strategy that regularizes cross-modal similarity according to sarcasm labels, encouraging genuine consistency in non-sarcastic samples while suppressing misleading consistency in sarcastic ones.
    
    \item Experiments on MMSD and MMSD2.0 verify the effectiveness of the proposed framework, and ablation studies further confirm the contribution of each component.
\end{itemize}

\section{Related Work}
Early sarcasm detection methods focused primarily on text-based approaches, but with the rise of social media, these methods proved insufficient. Schifanella et al.~\cite{Schifanella-2016-DetectingSarcasmMultimodal} were the first to combine text and visual information for sarcasm detection, followed by Cai et al.~\cite{Cai-2019-MultiModalSarcasmTwitter} who constructed the MMSD dataset and proposed a hierarchical fusion model. Later, Xu et al.~\cite{Xu-2020-ReasoningMultimodalSarcastic} and Pan et al.~\cite{Pan-2020-IntraInterIncongruity} introduced models like D\&R Net and Att-BERT that leveraged modality incongruity for sarcasm detection. Building on this, Liang et al.~\cite{Liang-2022-MultiModalCrossModalGCN} and Qiao et al.~\cite{Qiao-2023-MutualEnhancedIncongruity} further enhanced modality modeling using graph networks and additional visual cues such as OCR text. These studies improve cross-modal reasoning, but most of them still rely on predefined or insufficiently adaptive fusion schemes.

Recent work has focused on improving robustness and mitigating dataset bias in MSD. Qin et al.~\cite{Qin-2023-MMSD2} re-annotated MMSD to address false cues and class imbalance, and proposed Multiview-CLIP. Chen et al.~\cite{Chen-2024-InterCLIPMEP} further improved the Vision-Language interaction with an interactive encoder and memory-augmented predictor. Jia et al.~\cite{Jia-2024-DebiasingMultimodalSarcasm} explored debiasing via contrastive regularization, and Tang et al.~\cite{Tang-2024-LeveragingLLMsVisual} leveraged  retrieval to prompt LVLMs(e.g., LLaVA) for MSD. Wei et al.~\cite{Wei-2024-G2SAMGraphSemantic} proposed $G^2$SAM to integrate fine-grained multimodal graphs for stronger reasoning, while Yuan et al.~\cite{Yuan-2025-EnhancingSemanticAwareness} introduced ESAM by decoupling modalities and imposing sentimental congruity constraints with outlier masking. Although these methods improve robustness and multimodal reasoning, they still largely model sarcasm as generic cross-modal mismatch and do not explicitly address two challenges central to MSD: instance-dependent modality contribution and surface-level semantic consistency masking underlying intent contradiction.

\begin{figure*}
    \centering
    \includegraphics[width=0.75\linewidth]{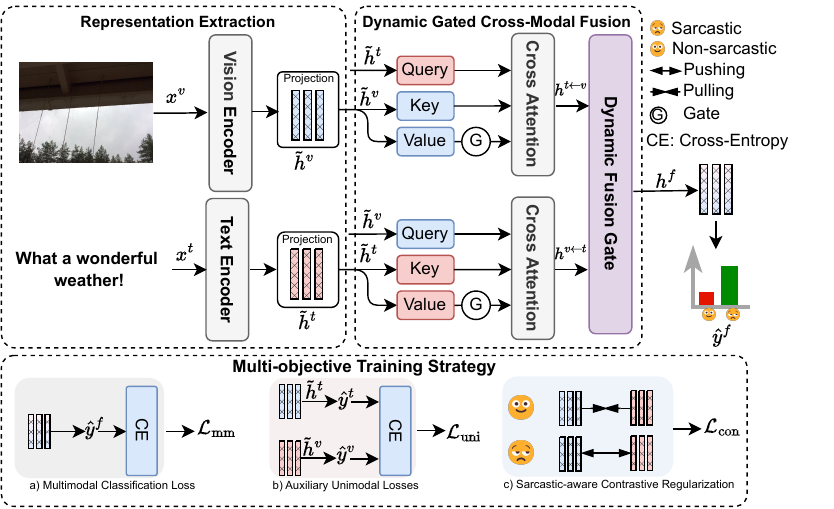}
    \caption{Illustration of the proposed framework for multimodal sarcasm detection}
    \label{fig:method}
\end{figure*}
\vspace{-5px}

\section{Methodology}
In this section, we introduce the proposed framework for multimodal sarcasm detection.
An overview of the architecture is illustrated in Fig.~\ref{fig:method}.
Given a multimodal input consisting of a text-image pair, our framework is organized as a unified architecture with multiple cooperative components.
First, modality-specific representations are extracted using a pre-trained vision-language encoder.
Then, a dynamic gated cross-modal fusion module performs bidirectional interactions between text and image and adaptively calibrates their contributions at the instance level.
To further reduce over-reliance on misleading cross-modal consistency, we incorporate a sarcastic-aware contrastive regularization objective during training, which applies label-aware constraints on cross-modal similarity.
Finally, the model is trained end-to-end with a multi-objective learning strategy that jointly optimizes the multimodal classification loss, auxiliary unimodal classification losses, and the sarcastic-aware contrastive regularization term.

\subsection{Representation Extraction}
Given a multimodal input consisting of text $x^t$ and image $x^v$, we employ a pretrained vision--language model, CLIP~\cite{Radford-2021-LearningTransferableVisual}, to extract representations. The parameters of both the text and vision encoders are fine-tuned during training.
Specifically, global textual and visual features are obtained from the text encoder $f_{}^{t}$ and vision encoder $f_{}^{v}$ of CLIP, respectively.
The extracted features $\tilde{h}^t$ and $\tilde{h}^v$ are then projected into a shared feature space via learnable linear transformations:
\begin{equation}
\tilde{h}^t = W_t f_{}^{t}(x^t), \quad
\tilde{h}^v = W_v f_{}^{v}(x^v).
\end{equation}

\subsection{Dynamic Gated Cross-Modal Fusion}
Given the projected textual and visual representations $\tilde{h}^t$ and $\tilde{h}^v$, we propose a dynamic gated cross-modal fusion module to capture bidirectional cross-modal interactions and adaptively balance modality contributions across instances.

\paragraph{Bidirectional Cross-Modal Attention}
Bidirectional cross-modal attention has been widely adopted to model interactions between textual and visual modalities.
However, most existing approaches aggregate attended cross-modal information indiscriminately, implicitly treating stronger semantic alignment as more reliable evidence.
Such an assumption may be problematic for multimodal sarcasm detection, where surface-level consistency between text and image can obscure the underlying contradictory intent.
Consequently, naively aggregating cross-modal information may emphasize misleading cues rather than truly informative sarcastic signals.

To address this issue, we introduce a value-gated bidirectional cross-modal attention mechanism, where learnable gates are applied to the value representations before cross-modal aggregation. 
Given the projected textual representation $\tilde{h}^t$ and visual representation $\tilde{h}^v$, we first compute gated value representations for each modality:
\begin{equation}
\bar{h}^v = \tilde{h}^v \odot \sigma(W_g^v \tilde{h}^v + b_g^v), \quad
\bar{h}^t = \tilde{h}^t \odot \sigma(W_g^t \tilde{h}^t + b_g^t),
\end{equation}
where $W_g^v$, $W_g^t$, $b_g^v$, and $b_g^t$ are learnable parameters, $\sigma(\cdot)$ denotes the sigmoid activation function, and $\odot$ denotes element-wise multiplication.

The gated representations are then used as value inputs in a symmetric bidirectional attention scheme:
\begin{equation}
h^{t \leftarrow v} = \mathrm{Attn}(\tilde{h}^t, \tilde{h}^v, \bar{h}^v), \quad
h^{v \leftarrow t} = \mathrm{Attn}(\tilde{h}^v, \tilde{h}^t, \bar{h}^t).
\end{equation}

Through this design, the model explicitly captures bidirectional semantic dependencies between modalities while selectively suppressing less informative or potentially misleading cross-modal signals at the value level. 
This is particularly beneficial for sarcasm detection, where superficial multimodal consistency may obscure the underlying incongruity.

\paragraph{Dynamic Fusion Gate}
Although value-gated attention effectively filters cross-modal information at a fine-grained level, the relative importance of textual and visual modalities still varies significantly across different instances.
In multimodal sarcasm detection, some samples are primarily driven by textual cues, while others rely more heavily on visual contradictions, making fixed fusion strategies suboptimal.

To address this challenge, we propose a dynamic fusion gate that adaptively balances the contributions of text-aware and image-aware representations in an instance-specific manner.

Given the text-aware visual representation $h^{t \leftarrow v}$ and the image-aware textual representation $h^{v \leftarrow t}$ obtained from gated bidirectional attention, we first concatenate them and compute a fusion gate $\boldsymbol{\alpha}$ as:
\begin{equation}
\boldsymbol{\alpha} = \sigma\left(W_f \left[ h^{t \leftarrow v} ; h^{v \leftarrow t} \right] + b_f \right),
\end{equation}
where $W_f$ and $b_f$ are learnable parameters and $\sigma(\cdot)$ denotes the sigmoid activation function.
The final fused representation $h^{f}$ is then obtained via gated combination:
\begin{equation}
h^{f} = \boldsymbol{\alpha} \odot h^{t \leftarrow v} + (1 - \boldsymbol{\alpha}) \odot h^{v \leftarrow t}.
\end{equation}
This dynamic fusion mechanism allows the model to selectively emphasize the modality that provides more discriminative evidence for each individual sample.

The fused representation $h^{f}$ is used as the final multimodal representation and serves as the sole input to the downstream sarcasm classifier during inference.

\subsection{Sarcastic-aware Contrastive Regularization}
While dynamic gated cross-modal fusion enables instance-adaptive integration of textual and visual cues, effective sarcasm detection further requires explicitly modeling the semantic inconsistency that characterizes sarcastic expressions.
In particular, sarcastic samples often exhibit surface-level semantic consistency between modalities while conveying contradictory underlying intent, which cannot be fully captured by classification objectives alone.

To this end, we introduce a sarcastic-aware contrastive regularization strategy as an auxiliary regularization objective.
Instead of uniformly encouraging cross-modal alignment, the proposed contrastive objective imposes label-aware constraints on the similarity between textual and visual representations, promoting semantic consistency for non-sarcastic samples while discouraging misleading consistency in sarcastic cases.

Formally, let $h^t = \mathrm{Norm}(\tilde{h}^t)$ and $h^v = \mathrm{Norm}(\tilde{h}^v)$ denote the $\ell_2$-normalized projected textual and visual representations extracted by the backbone encoder.
We compute the cosine similarity $s = \cos(h^t, h^v)$ and define a label-dependent contrastive loss as:
\begin{equation}
\mathcal{L}_{\text{con}} =
\begin{cases}
\max(0, m - s), & y = 0, \\
\max(0, s + m), & y = 1,
\end{cases}
\end{equation}
where $y \in \{0,1\}$ denotes the sarcasm label (0 for non-sarcastic and 1 for sarcastic), and $m>0$ is a predefined margin.
This formulation encourages higher cross-modal similarity for non-sarcastic samples while reducing over-reliance on potentially misleading cross-modal consistency in sarcastic ones, thereby helping the model distinguish genuine semantic alignment from sarcasm-related deceptive compatibility.

By incorporating sarcastic-aware contrastive regularization into the training process, the model learns more discriminative cross-modal representations that are sensitive to the unique semantic contradictions of sarcasm, complementing the proposed dynamic fusion mechanism.

\subsection{Multi-objective Training Strategy}
To effectively integrate dynamic gated cross-modal fusion and sarcastic-aware contrastive regularization, we adopt a unified multi-objective training strategy.
All components are optimized jointly in an end-to-end manner.

Given a training sample with label $y$, the overall training objective consists of three complementary loss terms:

\paragraph{Classification Loss}
The fused representation $h^{f}$, obtained via dynamic gated cross-modal fusion, is fed into a multimodal classifier.
We apply the standard cross-entropy loss $\mathrm{CE}$:
\begin{equation}
\mathcal{L}_{\text{mm}} = \mathrm{CE}(\hat{y}^{f}, y),
\end{equation}
where $\hat{y}^{f}$ denotes the predicted label distribution from the fused representation.

\paragraph{Auxiliary Unimodal Supervision}
To preserve modality-specific discriminative information and stabilize multimodal learning, we introduce auxiliary classification losses on the projected unimodal representations $\tilde{h}^t$ and $\tilde{h}^v$.
These unimodal classifiers share the same supervision label $y$ but are only used during training.
\begin{equation}
\mathcal{L}_{\text{uni}} = \mathrm{CE}(\hat{y}^{t}, y) + \mathrm{CE}(\hat{y}^{v}, y),
\end{equation}
where $\hat{y}^{t}$ and $\hat{y}^{v}$ are predictions from text-only and image-only classifiers, respectively.

\paragraph{Sarcastic-aware Contrastive Regularization}
Following the formulation described earlier, we apply a label-aware contrastive loss $\mathcal{L}_{\text{con}}$ between textual and visual representations.
The loss enforces high cross-modal similarity for non-sarcastic samples and penalizes excessive similarity for sarcastic ones, serving as a semantic inconsistency regularizer.

\paragraph{Overall Objective}
The final training objective is defined as:
\begin{equation}
\mathcal{L} = \mathcal{L}_{\text{mm}} + \mathcal{L}_{\text{uni}} + \mathcal{L}_{\text{con}}.
\end{equation}

This multi-objective formulation enables the model to jointly learn instance-adaptive cross-modal fusion, modality-aware representations, and sarcasm-sensitive semantic constraints.

\section{Experiments}
\subsection{Experiment Setting}
\paragraph{Datasets}
In this study, we evaluate our method on two widely used multimodal sarcasm detection benchmarks: MMSD~\cite{Cai-2019-MultiModalSarcasmTwitter} and its refined version MMSD2.0~\cite{Qin-2023-MMSD2}.
Both datasets consist of image--text pairs annotated with instance-level sarcasm labels, while MMSD2.0~\cite{Qin-2023-MMSD2} provides improved annotation quality and reduced noise.
The statistics of the two datasets are summarized in Table~\ref{tab:MMSD_Details}.
\begin{table}
\caption{Statistics of MMSD and MMSD2.0 datasets}
\centering
\label{tab:MMSD_Details}
\setlength{\tabcolsep}{2pt}
\begin{tabular}{l l r r r}
\toprule
\textbf{Dataset} & \textbf{Split} & \textbf{Total} & \textbf{Sarcastic} & \textbf{Non-sarcastic} \\
\midrule
\multirow{3}{*}{\textbf{MMSD}} 
& Train & 19,816 & 8,642 & 11,174 \\
& Val & 2,410 & 959 & 1,451 \\
& Test & 2,409 & 959 & 1,450 \\
\cmidrule(lr){1-5}
\multirow{3}{*}{\textbf{MMSD2.0}} 
& Train & 19,816 & 9,576 & 10,240 \\
& Val & 2,410 & 1,042 & 1,368 \\
& Test & 2,409 & 1,037 & 1,372 \\
\bottomrule
\end{tabular}
\end{table}

\paragraph{Baselines}
We compare our model with three types of baselines: 
(1) Text-modality models, including TextCNN~\cite{Kim-2014-CNNforSentenceClassification}, SMSD~\cite{Xiong-2019-SarcasmDetection}, and BERT~\cite{Devlin-2019-BERT}; 
(2) Image-modality models, including ResNet~\cite{He-2016-DeepResidualLearning} and ViT~\cite{Dosovitskiy-2020-ImageIsWorth}; 
(3) Multimodal models, including DIP~\cite{Wen-2023-DIPDualIncongruity}, Multi-view CLIP~\cite{Qin-2023-MMSD2}, MoBA~\cite{Xie-2024-MoBABiDirectionalAdapter}, G$^2$SAM~\cite{Wei-2024-G2SAMGraphSemantic}, TFCD~\cite{Zhu-2024-TFCDCounterfactualDebiasing}, DGLF~\cite{Zhu-2024-DGLF}, LLaVA+RAG~\cite{Tang-2024-LeveragingLLMsVisual}, and ESAM~\cite{Yuan-2025-EnhancingSemanticAwareness}. We additionally report the performance of GPT-5.4 under a zero-shot setting as a representative large multimodal model for reference. For a fair comparison, all methods are evaluated using the same data splits and evaluation metrics as previous work.

\subsection{Main Results}
\begin{table*}[t]
\caption{Performance of selected baselines and our method(\%).}
\label{tab:selected_baselines}
\centering
\small
\setlength{\tabcolsep}{4pt}
\begin{tabular}{llcccc|cccc}
\toprule
\textbf{Modality} & \textbf{Method} & \multicolumn{4}{c}{\textbf{MMSD}} & \multicolumn{4}{c}{\textbf{MMSD2.0}} \\
\cmidrule(lr){3-6} \cmidrule(lr){7-10}
 &  & Acc.\% & P\% & R\% & F1\% & Acc.\% & P\% & R\% & F1\% \\
\midrule
\multirow{3}{*}{\textbf{Text}} 
& TextCNN & 80.03 & 74.29 & 76.39 & 75.32 & 71.61 & 64.62 & 75.22 & 69.52 \\
& SMSD    & 80.90 & 76.46 & 75.18 & 75.82 & 73.56 & 68.45 & 71.55 & 69.97 \\
& BERT & 83.60 & 78.50 & 82.51 & 80.45 & 76.52 & 74.48 & 73.09 & 73.91 \\
\midrule
\multirow{2}{*}{\textbf{Image}} 
& ResNet & 64.76 & 54.41 & 70.80 & 61.53 & 65.50 & 61.17 & 54.39 & 57.58 \\
& ViT    & 67.83 & 57.93 & 70.07 & 63.40 & 72.02 & 65.26 & 74.83 & 69.72 \\
\midrule
\multirow{8}{*}{\textbf{Multimodal}} 
& DIP                     & 89.59 & 87.76 & 86.58 & 87.17 & 80.96 & 78.02 & 77.56 & 77.79 \\
& Multi-view CLIP         & 88.33 & 82.66 & 88.65 & 85.55 & 85.64 & 80.33 & 88.24 & 84.10 \\
& MoBA                    & 88.96 & 82.84 & 88.12 & 85.40 & 85.83 & 80.42 & 88.67 & 84.34 \\
& G$^2$SAM                & 90.48 & 87.95 & 89.02 & 88.48 & 79.43 & 72.04 & 78.07 & 78.07 \\
& TFCD                    & 89.57 & 84.83 & 89.43 & 88.13 & 86.54 & 82.46 & 87.95 & 84.31 \\
& DGLF                    & 89.43 & 85.81 & 89.27 & 87.51 & 86.82 & 81.90 & \textbf{89.85} & 85.69 \\
& LLaVA+RAG               & 89.97 & 89.26 & 89.58 & 89.42 & 86.43 & 87.00 & 86.30 & 86.34 \\
& ESAM                    & 90.11 & 86.87 & 89.54 & 88.19 & 85.87 & 83.12 & 86.05 & 84.56 \\
& GPT-5.4 (zeroshot)                 & 71.05 & 76.51 & 75.50 & 71.01 & 72.85 & 78.79 & 75.78 & 72.55 \\
\midrule
& \textbf{Ours}          & \textbf{92.62} & \textbf{91.96} & \textbf{92.82} & \textbf{92.33} & \textbf{89.66} & \textbf{89.36} & 89.74 & \textbf{89.51} \\
\bottomrule
\end{tabular}
\end{table*}
As shown in Table~\ref{tab:selected_baselines}, our method achieves competitive or superior performance across most evaluation metrics on both MMSD and MMSD2.0.
On the MMSD benchmark, our model achieves the best F1 score, surpassing the strongest baseline by a clear margin.
Notably, similar performance gains are observed on MMSD2.0, which is an improved version of MMSD with refined re-annotations and reduced spurious cues, indicating that the improvements are not limited to the original dataset.
Compared with recent CLIP-based and large vision--language models, our approach demonstrates superior effectiveness, highlighting the limitation of fixed cross-modal alignment strategies for sarcasm detection.
We attribute these gains to the proposed instance-adaptive fusion mechanism, together with the sarcastic-aware contrastive regularization that explicitly suppresses misleading semantic consistency between modalities.

\subsection{Ablation Study}
We perform ablations under the same backbone, training hyper-parameters, and evaluation protocol, and only change the specified module/loss. We compare Full with the following variants: (1) w/o CMI, which removes cross-modal interaction and uses fixed fusion of unimodal features; (2) w/o BiXAtt, which removes one cross-attention direction; (3) w/o VGate, which disables value gating in cross-attention; (4) w/o DFGate, which replaces the dynamic fusion gate with a fixed fusion strategy; (5) w/o SaCR, which removes sarcastic-aware contrastive regularization; and (6) w/o UniAux, which removes unimodal auxiliary classification supervision. Performance is reported with the same metrics as the main results.

\begin{table}[t]
\caption{Experiment results of ablation study (\%).}
\label{tab:ablation_study}
\centering
\begin{tabular}{@{}lcccc@{}}
\toprule
\textbf{Variant} & \multicolumn{2}{c}{\textbf{MMSD}} & \multicolumn{2}{c}{\textbf{MMSD2.0}} \\
\cmidrule(lr){2-3} \cmidrule(lr){4-5}
& Acc.\% & F1\% & Acc.\% & F1\% \\
\midrule
Full
& \textbf{92.62} & \textbf{92.33}
& \textbf{89.66} & \textbf{89.36} \\
\midrule
w/o CMI
& 87.91 & 87.42
& 85.10 & 84.97 \\
w/o BiXAtt (only v$\rightarrow$t)
& 87.48 & 87.03
& 81.86 & 81.70 \\
w/o BiXAtt (only t$\rightarrow$v)
& 81.84 & 81.07
& 80.53 & 80.37 \\
w/o VGate
& 92.08 & 91.77
& 88.71 & 88.59 \\
w/o DFGate
& 90.35 & 89.80
& 88.34 & 88.28 \\
w/o SaCR
& 91.28 & 91.00
& 88.54 & 88.49 \\
w/o UniAux
& 91.31 & 90.92
& 89.16 & 89.03 \\
\bottomrule
\end{tabular}
\end{table}
Table~\ref{tab:ablation_study} shows that removing any component consistently degrades performance on both MMSD and MMSD2.0. In particular, removing cross-modal interaction (w/o CMI) causes a large drop in F1 (5.40\% on MMSD and 4.54\% on MMSD2.0), highlighting the necessity of explicit cross-modal reasoning for sarcasm-related incongruity. Disabling bidirectionality further hurts performance, and the only t$\rightarrow$v variant yields the lowest scores (F1 81.07/80.37\%), suggesting that symmetric text–image interactions are important. w/o VGate reduces F1 by 1.05/0.92\%, indicating that value-level gating helps suppress misleading or redundant attended cues. Replacing the dynamic fusion gate with fixed fusion (w/o DFGate) results in a clear degradation (3.02/1.23\%), validating instance-adaptive modality balancing. Removing SaCR also lowers F1 (1.82/1.02\%), showing the benefit of sarcasm-aware contrastive regularization. Finally, removing UniAux also degrades performance, suggesting that auxiliary unimodal supervision provides additional training signals for representation learning. Overall, cross-modal interaction and dynamic fusion contribute the most, and the consistent trends on MMSD2.0 confirm robustness.

\subsection{Case Study and Visualization}
\begin{figure}
    \centering
    \includegraphics[width=0.85\linewidth]{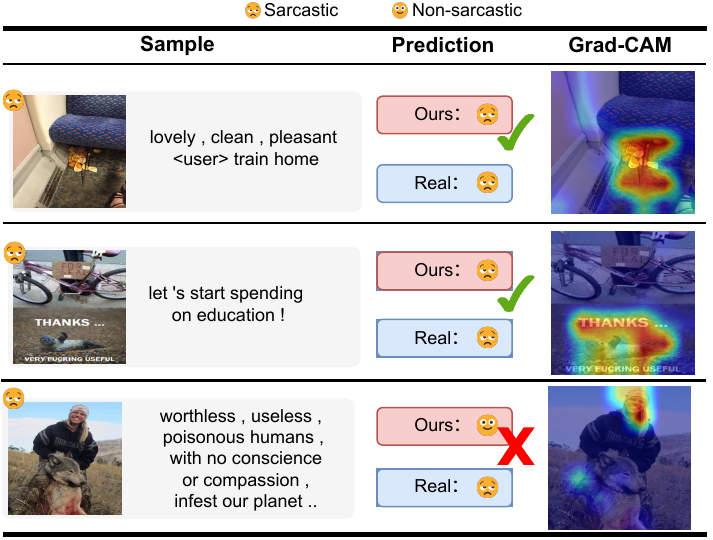}
    \caption{Example of case study and visualization.}
    \label{fig:case-study}
\end{figure}
Fig.~\ref{fig:case-study} presents two correctly classified sarcastic samples and one failure case, together with Grad-CAM visualizations on the image encoder.
In the first example, the text describes a ``lovely, clean, pleasant'' train ride, but the image shows obvious litter on the floor, and Grad-CAM highlights the litter region as the key evidence for sarcasm recognition. 
In the second example, the text expresses a positive statement (``let's start spending on education!''), while the meme image contains an explicit ironic cue (e.g., ``THANKS ... VERY ... USEFUL''), and Grad-CAM focuses on the meme text area, which supports the sarcastic prediction. 
This failure suggests a human-centric saliency bias: the text mentions ``humans'', and the Grad-CAM map concentrates on the face region, while the model under-attends other cues that better support the intended irony.
These cases show that our model captures typical sarcasm cues, but it can over-rely on cross-modal alignment in some scenarios and overlook the true evidence of sarcasm.

\section{Conclusion}
In this paper, we address multimodal sarcasm detection from the perspective of two key challenges: instance-dependent modality contribution and misleading surface-level cross-modal consistency. To this end, we propose a unified framework that combines dynamic gated cross-modal fusion with sarcastic-aware contrastive regularization. The proposed model performs bidirectional text-image interaction, adaptively balances multimodal evidence for each instance, and applies label-aware regularization on cross-modal similarity during training. Experiments on MMSD and MMSD2.0 show the effectiveness of the proposed framework, while ablation and case studies further illustrate the role of each component. In future work, we will explore finer-grained visual-textual grounding and more robust sarcasm modeling strategies that are less sensitive to superficial alignment cues.

\bibliographystyle{IEEEtran}
\bibliography{ref}

\end{document}